\documentclass[runningheads]{llncs}
\usepackage{booktabs} 
\usepackage{graphicx}
\usepackage{times}
\usepackage{latexsym}
\usepackage{tabularx}
\usepackage{graphicx}
\usepackage{amsmath}
\usepackage{hyperref}
\usepackage{multirow}
\usepackage[vietnamese]{babel}
\usepackage{placeins}
\usepackage[referable]{threeparttablex}
\usepackage{xcolor}
\usepackage{tablefootnote}
\newcolumntype{P}[1]{>{\centering\arraybackslash}p{#1}}
\begin{document}
\renewcommand{\tablename}{Table}
\renewcommand{\figurename}{Figure}
\renewcommand{\refname}{References}
\renewcommand{\abstractname}{Abstract}

\title{A Combination of BERT and Transformer \\ for Vietnamese Spelling Correction}

\titlerunning{BERT \& Transformer for Vietnamese Spelling Correction}

\author{Hieu Ngo Trung\inst{1}, Duong Tran Ham\inst{1}, Tin Huynh\inst{1}, Kiem Hoang \inst{1}}
\authorrunning{Hieu et al.}

\institute{The Saigon International University, Vietnam \\
\email{ \{ngotrunghieu,tranhamduong,huynhngoctin,hoangkiem\}@siu.edu.vn }  }

\maketitle              
\begin{abstract}
Recently, many studies have shown the efficiency of using \textbf{B}idirectional \textbf{E}ncoder \textbf{R}epresentations from \textbf{T}ransformers (BERT) in various Natural Language Processing (NLP) tasks. Specifically, English spelling correction task that uses Encoder-Decoder architecture and takes advantage of BERT has achieved state-of-the-art result. However, to our knowledge, there is no implementation in Vietnamese yet. Therefore, in this study, a combination of Transformer architecture (state-of-the-art for Encoder-Decoder model) and BERT was proposed to deal with Vietnamese spelling correction. The experiment results have shown that our model  outperforms other approaches as well as the Google Docs Spell Checking tool, achieves an 86.24 BLEU score on this task.

\end{abstract}
\section{Introduction}
A spelling error is a word written in a wrong spelling standard, including various forms: homophone, acronym, uppercase, or fairly the phenomenon of wrong-written words. Usually, there are many groups of origination causing the spelling errors to happen in Vietnamese: typing, semantic confusion, local pronunciation, rules, and standards in the written text, not mastery in grammar and influence of social network language, etc \cite{PhanXuan2017}.

Spelling correction is a Natural Language Processing task that focuses on correcting spelling errors in text or a document. The spelling correction task keeps a critical role in enhancing the user typing experience and guarantees the information integrity of Vietnamese. Besides, one primary application is the ability to incorporate with other tasks. For example, when using the spelling correction attached to the last phase of the Scene Text Detection / Optical Character Recognition (OCR) task, the results are improved significantly \cite{ocr-error-correction,ocr-error-correction-class}. Consider the chatbot task, if spelling correction is applied to preprocess user inputs, the chatbot will have better accuracy and performance in understanding the user requests \cite{error-correction-chatbot}.

Frequently, the spelling correction can be divided into two steps, including spell checking and spell correcting \cite{survey-error-correction}. In the first phase, mistakes are investigated if there are any in the given input and then try to transform the wrong words into corrected words in the second phase.

Contrary to English and other languages, the Vietnamese possess up to six complex diacritic marks and uses them as a discrimination sign. Therefore, a word that combines with different diacritic marks can create up to six written forms, and each of them also has independent meaning and usage. For instance, the word “ma” (ghost) can be written in 5 more ways with 5 different diacritic marks: “má” (mother), “mà” (nevertheless), “mả" (tomb), “mã" (code), “mạ" (rice seedlings). All the originations and elements described above have made the Vietnamese spelling correction problem a very challenging task.

There are many initial approaches to the Vietnamese Spelling Correction task that has been carried out such as applying rule-based methods \cite{10.3115/1075218.1075294}, using edit-distance algorithm \cite{edit-distance-error-correction}, collating with dictionaries,  using n-gram/big-gram language model \cite{n-gram-vietnamese-spellchecking}, etc. However, most of these approaches neither adapted to out-of-vocabulary words nor did they take the contextualized word embeddings into account. In order to deal with these gaps, many deep learning models using Recurrent Neural Network (RNN) or Long Short-term Memory (LSTM) networks have been proposed and achieved impressive performance \cite{error-correction-blstm}.

Recently, spelling correction studies that took advantage of the Encoder-Decoder model have attracted much attention and achieved state-of-the-art in the English spelling correction task \cite{liu-etal-2018-automatic,error-correction-nmt}. This is a novel approach, which is notably potential because of the optimal utilization of the parallelism calculation ability and the strength of powerful pre-trained language models. One of the most attention is the usage of the Transformer architecture \cite{vaswani2017attention} with the language model known as BERT \cite{DBLP:journals/corr/abs-1810-04805}. Despite its success in English \cite{zhu2020incorporating}, there is still no implementation in Vietnamese that can be used in practice. Therefore, this paper aims to apply these architectures and techniques to improve the performance of correcting Vietnamese spelling errors. The experiment results show that the proposed solution achieves considerable efficiency and is able to integrate with practical services. The main contributions of this study could be summarized as follows:
\begin{itemize}
\item Applying the Transformer architecture and leveraging the pre-trained BERT to provide a solution to the Vietnamese spelling correction problem.
\item Constructing a large and creditable dataset based on the most common practical Vietnamese spelling errors. The evaluation dataset is published for the Vietnamese NLP community using in related works.
\end{itemize}

The remainder of this paper is organized as follows. In section \ref{relatedwork}, related works are presented and discussed. Section \ref{OurApproach} describes the proposed methods in detail. The dataset, experimental results, and discussion are provided in section \ref{EXPERIMENTS}. Section \ref{Conclusion} summarizes, concludes and gives future orientation.

\section{Related Works}\label{relatedwork}
Spelling correction is not a new problem in NLP tasks. Earlier there have been many approaches for this problem, from straightforward approach using probability, such as implementing the Naive Bayes algorithm (Peter Norvig\footnote{https://norvig.com/spell-correct.html}). The large N-gam-based language modeling approach of both left and right side has improved the performance of spelling correction tasks \cite{n-gram-vietnamese-spellchecking}. After training with a large corpus, this model can predict the probabilities of multiple N-gram candidates for correcting words. Large N-gram LM is a pure probability approach. It expects high memory resources to store all pre-calculated probabilities of N-gram pairs and can not handle a not-pre-trained error, which leads to all of the probability of N-gram pairs to zero.

The advantages of contextual embedding in word presentation model, likes Word2Vec \cite{mikolov2013efficient}, Glove \cite{pennington-etal-2014-glove}, etc, is being taken into the spelling correction task \cite{fivez-etal-2017-unsupervised}. An edit-distance algorithm generates the candidates, then each candidate's score is calculated by the cosine similarity between the candidate vector and the target word vector, the highest score ranking candidate will be selected. This method has shown significant results in the spelling correction task and is suitable for many languages, especially in Vietnamese, English, etc. On the other hand, this approach requires many resources to represent the rich context embeddings of a language accurately. Also, out-of-vocabulary (OOV) is a large major problem to the ranking system.

Another approach to using deep learning has been developed through the use of LSTM network \cite{DBLP:conf/pacling/NguyenDN19}. A LSTM network \cite{lstm} is constructed that encodes the input sequence and then decodes it to the expected correct output sequence, respectively. The accuracy of their model makes a significant gap compared to the current state-of-the-art model \cite{n-gram-vietnamese-spellchecking}. Studies have reported that spelling correction can be beneficial from Encoder-Decoder architecture \cite{kaneko2020encoderdecoder,pham2019grammatical}. A state-of-the-art approach in English is implemented by the Encoder-Decoder architecture \cite{10.5555/2969033.2969173} and also makes use of the powerful pretrained BERT model \cite{kaneko2020encoderdecoder}. They first fine-tuned the BERT model and then used its last hidden presentation output as additional features to the an error correction model, which is a customized Transformers \cite{vaswani2017attention} architecture. A similar method for Vietnamese grammatical error correction using OpenNMT framework \cite{klein-etal-2017-opennmt} instead of the Transformer architecture \cite{pham2019grammatical}. This method, respectively, depends on using the Microsoft Office spelling tool to check and detect the incorrect text before the correction step.

Through previous related works, deep learning approaches to spelling correction are our focus. The approach is receiving much concern, gains state-of-the-art performance is the Encoder-Decoder architecture with prominent pre-trained MLM. Both well-known pre-trained Google Multilingual \cite{DBLP:journals/corr/abs-1810-04805} and vinai/phobert \cite{nguyen2020phobert} are used to extract hidden presentations and implement the transformer architecture into a specific Vietnamese spelling correction task.

\section{Our approach}\label{OurApproach}
\label{sec:length}

\subsection{Introduction to the Vietnamese language}
This section briefly presents the characteristics and differences from English of the Vietnamese language. Unlike neighbor countries, the Vietnamese does not use hieroglyphic letters, but a modified Latin (Roman) alphabet. The Vietnamese alphabet uses 29 letters, unlike the English alphabet, it does not use 4 letters 'w', 'f', 'j', 'z' and uses 6 more vowel letters (with special mark): 'ă', 'â', 'ê', 'ô', 'ơ', 'ê', and the letter 'đ' \cite{vietnamese-textbook-grade01,vietnamese-vietnamese-lecture-vietnamese-people}. Along with the above 6 types of diacritics, it forms up to 67 separate letters (nearly 3 times larger than the number of letters in English). Therefore, spelling mistakes are much more common in Vietnamese than in English.

\subsection{Analyzing of Vietnamese common spelling error}
In this section, the concept of common error type in the Vietnamese language are presented. Due to the lacks of scientific public research or national survey constructed on this topic, various types of Vietnamese error type from previous related work \cite{DBLP:conf/pacling/NguyenDN19,n-gram-vietnamese-spellchecking,edit-distance-error-correction} are summarized and divided them into six groups:

\begin{itemize}
    \item \textbf{Abbreviation: }There are a wide variety of abbreviation for common words in Vietnamese writing. Despite its convenience, this style of writing may raise misunderstanding, make the writing less formal and not accepted by most people. To determine this error cases, a list of most common abbreviation substitutions in Vietnamese is compiled from the Internet.
    
    \item \textbf{Region:} The region error type is the most complicated type to analyze owing to its variety of happening contexts. The region error type comes from different region pronunciation across the Vietnam territory. When people tend to write a word the same way they pronounce it, this error occurs. Many adults may mistake this type of error if not a native speaker or do not have enough knowledge of the Vietnamese language. An incorrect word with region type stands alone, may still have meaning. Some examples of region error type are described in table \ref{table:RegionErrorTypes}.
    
    \begin{table}[h!]
    {\centering
    \caption{Some examples of region error type}    
        \begin{tabular}{|P{2.5cm}|P{2.5cm}|P{2.5cm}|P{2.5cm}|}
        \hline
        Original & Usually mistake for & Original & Usually mistake for \\
        
        \hline
        ch- & tr- & -nh & -n \\
        tr- & ch & c- & k- \\
        -n & -ng & k- & c- \\
        -ng & -n & ngh- & ng- \\
        g- & gi- & gi- & g- \\ 
        ... & ... & ... & ... \\
        \hline
        \end{tabular}
        \label{table:RegionErrorTypes}
    \par}
    \end{table}

        
    
    \item \textbf{Teencode:} Teencode (or Teen-code) is a method of writing used by teenagers on social media or through messaging. Those teenagers put words into special encryption so the adults can not understand.
    
    \item \textbf{Telex:} Telex is a convention for encoding Vietnamese text in plain ASCII characters, used initially for transmitting Vietnamese text over telex systems. Forgetting to turn on the language encoder or entering the wrong Vietnamese Telex rules leads to this type of error.
    
    \item \textbf{Fat Finger} Fat Finger, also known as the clumsy finger, means when typing through a cell phone or computer keyboard, the user's finger mistypes the surrounding key instead of the target key, causing the wrong words.
    
    \item \textbf{Edit Distance} Edit Distance is a pseudo error generation strategy in which several characters equal to a 'distance' to the original are randomly replaces. Although this error rarely happens logically, a low percentage amount is still generated in our data set.
\end{itemize}

For the convenience of observation, a list of examples corresponding to the type of error is presented in table \ref{tab:dataset-summarization}.

\begin{table}[h!]
    \caption{A summary about error types}
    \label{tab:dataset-summarization}
\begin{tabular}{|c|l|ll|}
\hline
\multirow{2}{*}{\textbf{Error type}} & \multicolumn{1}{c|}{\multirow{2}{*}{\textbf{When it happends}}} & \multicolumn{2}{c|}{\textbf{Examples}} \\ \cline{3-4} 
 & \multicolumn{1}{c|}{} & \multicolumn{1}{c|}{\textbf{Correct}} & \multicolumn{1}{c|}{\textbf{Incorrect}} \\ \hline
Abbreviation & \begin{tabular}[c]{@{}l@{}}To make writing faster and more conv-\\ enient, people use abbreviation instead of \\ full words.\end{tabular} & \multicolumn{1}{l|}{\begin{tabular}[c]{@{}l@{}}Không (No)\\ Mọi người (Everyone)\\ Bình thường (Normal)\end{tabular}} & \begin{tabular}[c]{@{}l@{}}\textbf{Kg}\\ \textbf{Mn}\\ \textbf{Bt}\end{tabular} \\ \hline
TeenTeencode & \begin{tabular}[c]{@{}l@{}}When teenagers try to encode their text in\\  order to hide information.\end{tabular} & \multicolumn{1}{l|}{\begin{tabular}[c]{@{}l@{}}Ví dụ (Example)\\ Chồng (Husband)\\ Điện thoại (Phone)\end{tabular}} & \begin{tabular}[c]{@{}l@{}}V\textbf{j} d\textbf{u}\\ \textbf{Cho`ng}\\ \textbf{Dj3n tk04j}\end{tabular} \\ \hline
Fat-Finger & \begin{tabular}[c]{@{}l@{}}While typing with a virtual keyboard, the\\ user’s finger mistypes the surroundings in-\\ stead of the target key.\end{tabular} & \multicolumn{1}{l|}{\begin{tabular}[c]{@{}l@{}}Xin chào (Hello)\\ Trường học (School)\\ Điện thoại (Phone)\end{tabular}} & \begin{tabular}[c]{@{}l@{}}X\textbf{im} chào\\ Trường h\textbf{ịc}\\ Điện th\textbf{ọak}\end{tabular} \\ \hline
Telex & \begin{tabular}[c]{@{}l@{}}Forgetting to turn on the language encoder\\ or entering the wrong Vietnamese Telex\\ rules.\end{tabular} & \multicolumn{1}{l|}{\begin{tabular}[c]{@{}l@{}}Xin chào (Hello) \\ Trường học (School) \\ Điện thoại (Phone)\end{tabular}} & \begin{tabular}[c]{@{}l@{}}Xin ch\textbf{afo}\\ Tr\textbf{uowng} h\textbf{ojc}\\ \textbf{Ddieejn thoaji}\end{tabular} \\ \hline
Region & \begin{tabular}[c]{@{}l@{}}Different region pronunciation across the\\ Vietnam (as people tend to write the same\\ as as they pronounce).\end{tabular} & \multicolumn{1}{l|}{\begin{tabular}[c]{@{}l@{}}Tranh (Painting)\\ Lạnh (Cold) \\ Nghỉ (Rest)\end{tabular}} & \begin{tabular}[c]{@{}l@{}}\textbf{Ch}anh (Lemon)\\ \textbf{N}ạnh\\ Ngh\textbf{ĩ}(Think)\end{tabular} \\ \hline
\multicolumn{1}{|l|}{Edit-Distance} & \begin{tabular}[c]{@{}l@{}}A common pseudo misspelling error gen-\\ eration strategy.\end{tabular} & \multicolumn{1}{l|}{\begin{tabular}[c]{@{}l@{}}Tranh (Painting) \\ Lạnh (Cold) \\ Mưa (Rain)\end{tabular}} & \begin{tabular}[c]{@{}l@{}}\textbf{Th}anh (Bar)\\ L\textbf{h}oạnh\\ M\textbf{t}ưa\end{tabular} \\ \hline
\end{tabular}
\end{table}

\subsection{BERT} \label{subsection:BERT}
BERT is a language representation model based on multi-layer bidirectional transformers encoder architecture. There is a wide variety of challenging natural language tasks that BERT can handle and achieve state-of-the-art performance, from classification, question answering, and sequence-to-sequence learning task, etc. BERT can represent sentence effectively by its encoding mechanism, including various embedding step as token embeddings, sentence embeddings, and transformers positional embeddings. Then, this BERT's last hidden presentation output from BERT is used as the input into the transformers architecture. 

In this study, two well pre-trained BERT on the Vietnamese: the Google Multilingual BERT\footnote{Github: https://github.com/google-research/bert} (bert-base-multilingual-cased) and VinAI/phoBERT\footnote{Github: https://github.com/VinAIResearch/PhoBERT} are considered. They are both trained with extensive Vietnamese corpus, while the multilingual BERT is the BERT base model, and phoBERT is a RoBERTa model \cite{Liu2019RoBERTaAR} (which is a modified version from the base model).


\subsection{Transformers}

Before the transformers architecture, Encoder-Decoder architecture using RNN/ LSTM/ GRU (Gated Recurrent Unit) cell is used widely in machine translation and sequence-to-sequence tasks. This Encoder-Decoder architecture, also known as the Seq2Seq model, uses several RNN cells to encode the input tokens to hidden states and then sum all hidden states up before sending them to the decoder. Thanks to this hidden state, the decoders receive all previously encoded information and use it for the output token prediction task. Despite large capacities in handling sequence-to-sequence tasks, the Seq2Seq decoder may fail to fully capture the meaning and context of the last hidden presentation from the encoder. That means the more extended and more complex the input sequence, the less effective the hidden presentation can represent, which is known as the bottleneck problem.

While the attention mechanism takes several inputs simultaneously, construct weight matrices captured from each hidden presentation input to calculate a weighted sum of all the past encoder states. The decoder will then take the inputs and the provided attention weights, and through that, the decoder knows how to 'pay much attention' to which hidden presentation and vice versa. Another limitation of Seq2Seq architecture is that it handles the input sequentially, which means to compute for the current token at time \textit{t}, we need the previous hidden state \textit{t-1} and so on. Therefore, especially in the spelling correction task, if many erroneous tokens stick together, the correction of the last tokens can be poorly affected. 
The transformer architecture and attention mechanism come to cross those boundaries of those previous architectures. Based on the Encoder-Decoder architecture \cite{10.5555/2969033.2969173}, the transformers architecture \cite{vaswani2017attention} uses stacked multi-head self-attention and fully connected layers. The transformer is designed to allow parallel computation and reduce the drop in performance due to long dependencies. It uses positional embeddings and multi-head self-attention to encode more information about the position of a token and the relation between each token.

As Shun has provided an advanced insights to incorporating pseudo data into the spelling correction task \cite{DBLP:journals/corr/abs-1909-00502}. Consequently, a vast pseudo training dataset can be generated so that not only can the transformer maximize its parallelization ability but also BERT can represent its rich contextual embedding vectors.

\subsection{Incorporate BERT into Transformers}

As mentioned in section \ref{subsection:BERT}, BERT is capable of deep language understanding by capturing contextual embedding of different words in a sequence. In addition, the Transformer model has been proved to be more efficient than popular Encoder-Decoder architectures, especially in the Machine Translation problem. 

\begin{figure}[h]
\centerline{\includegraphics[width=7.3cm]{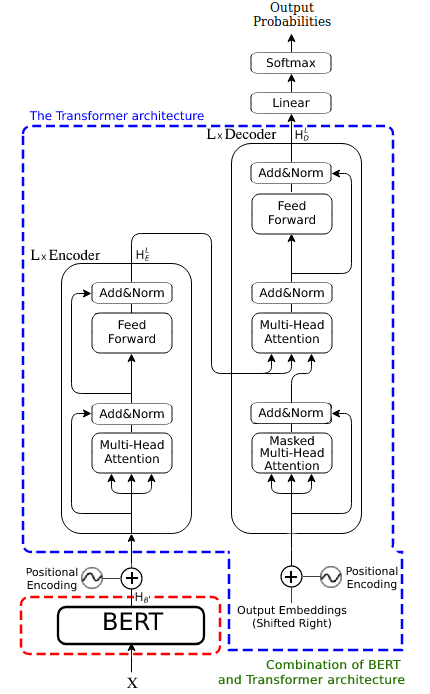}}
\caption{Proposed combination between BERT and Transformer}
\label{fig:transformer}
\end{figure}

Some recent studies have treated the spelling correction problem as a machine translation job where the error sentence is the source sequence and the corrected sentence is the target sequence. And the combination of BERT and Transformer achieves 
state-of-the-art results for the English spelling correction task \cite{kaneko2020encoderdecoder}. However, to the best of our knowledge, there has not been any research combining BERT and Transformer for Vietnamese spelling correction problem. The combination can be briefly summarized in the following steps:

\begin{itemize}
    \item {\bf Step 1}: Let the input sentence notated as X = (\(x_1,...,x_n)\), where \(n\) is the number of its tokens; \(x_i\) is the \textit{i}-th token in X. BERT receives the input sequence tokens, and through its layers, BERT extracts them to hidden presentations notated as \textit{\(H_B\)} = (\(h_1, h_2, ..., h_n)\), where {\(H_B\)} is the output of the last layers in BERT.

\item {\bf Step 2}:  The Encoder will take {\(H_B\)} from the previous step and encode the representation of each l layer {\(H_E^l\)}. The final contextual representation of the last encoder layer {\(H_E^L\)} is the output of the Encoder. The Encoder components consist of the multi-head self-attention mechanism, position-wise fully connected feed-forward network. A residual connection around each of the two sub-layers, followed by layer normalization. A Multi-head Attention is a component allowing the model to jointly attend to information from different representations and helps the encoder look at other words in the input sentence as it encodes a specific word for better-capturing contextual embedding. 

\item {\bf Step 3}: The Decoder receives the representation {\(H_E^L\)} from the Encoder and decodes through its layers into final representation {\(H_D^L\)}. Similar to the Encoder, the Decoder possessed the same components of the Encoder. These Encoder {\(H_E^L\)} are to be used by each Decoder in its “encoder-decoder attention” layer which helps the Decoder focus on appropriate places in the input sequence. 

\item {\bf Step 4}: Finally, the Decoder final representation {\(H_D^L\)} is mapped via a linear transformation and softmax to get the \textit{t}-th predicted word \^{y}. The decoding process continues until meeting the end-of-sentence token.

\end{itemize}
An illustration of our proposed method is shown in Figure~\ref{fig:transformer}.

\section{Experimental Evaluation} \label{EXPERIMENTS}
This section includes dataset description, evaluation method, model hyper-parameter setting as well as experimental results of applying the Transformer architecture and BERT to Vietnamese spelling correction.
\subsection{Experimental dataset}
This section describes the process of creating our training and testing set based on the Binhvq News Corpus\footnote{Github: https://github.com/binhvq/news-corpus} which contains 14,896,998 Vietnamese news crawled from the Internet and preprocessed, including steps like HTML tag removal, duplicate removal, NFC standardization, and sentence segmentation. The corpus is gathered from reputable news and media sites in Vietnam, so the data is very reliable in terms of spelling. For the purpose of training and evaluating spelling correction, our newly constructed dataset must consist of two fields that can be described as a pair of correct and incorrect spelling sentences.  

To the best of our knowledge, there is no specific survey as well as assessment on the rate of error types appearing in Vietnamese. However, Vietnamese often has common spelling mistakes: Region, FatFinger, Telex. Besides, some other types of errors are concerned, such as Edit-Distance, Abbreviation, Teencode, but rarely happened in practice. Therefore, error rate is reproduced based on our experience. Details of the rates of error types in the generated data set are listed in table \ref{table:typeOfError}.

\FloatBarrier
\begin{table}[h]
	\centering
	\caption{Training and Testing sets}
\begin{tabularx}{\columnwidth}{|X|c|}
\hline
\textbf{Error Type} & \textbf{Error Ratio (\%)}   \\
\hline
Acronym & 3.0 \\
\hline
Teencode sets & 3.0\\
\hline
Edit-Distance & 3.0\\
\hline
FatFinger & 30.0\\
\hline
Telex & 30.0\\
\hline
Region & 31.0\\
\hline

\hline
\end{tabularx}
\label{table:typeOfError}
\end{table}
\FloatBarrier

The training set is composed by randomly selecting 4,000,000 sentences from the Binhvq corpus and then apply the pseudo error generator to these correct ones. All the sentences must have an average word count between 50-60 words per sentence. Similarly, validating set and testing set are generated with the number of correct sentences chosen from the above corpus 20,000 and 6,000 respectively. Details of the dataset can be summarized in table \ref{table:dataset}. The testing set is public and can be downloaded at the following link \footnote{Github: https://github.com/tranhamduong/Vietnamese-Spelling-Correction-testset}.

\FloatBarrier
\begin{table}[h]
	\centering
	\caption{Training and Testing sets}
\begin{tabularx}{\columnwidth}{|l|X|X|}
\hline
\textbf{Dataset} & \textbf{Size (\#Pair of sentence)} & \textbf{Avg. Length per sentence (\#token)}  \\
\hline
Training sets & 4,000,000 & 60 \\
\hline
Validating sets & 20,000 &  60\\
\hline
Testing sets & 6,000 & 60\\
\hline
\end{tabularx}
\label{table:dataset}
\end{table}
\FloatBarrier

\subsection{Evaluating Metric}
In the perspective of a spelling correction task, many traditional approaches used Accuracy, Precision, Recall, and F1 for evaluation \cite{DBLP:conf/pacling/NguyenDN19,n-gram-vietnamese-spellchecking}. These metrics require the predictions' words to have the same length as labels' words. Recently, BLEU score is chosen, especially in deep learning models because of its ability to adapt to different prediction lengths \cite{kaneko2020encoderdecoder,zhu2020incorporating}. Therefore, BLUE \cite{papineni2002bleu} is selected for the evaluating task. BLEU, or the Bilingual Evaluation Understudy, is a score for comparing a candidate translation of text to one or more reference translations. Although developed for translation, it can be used to evaluate text generated for a suite of natural language processing tasks. Our BLEU configuration uses four n-grams settings because the spelling correction task critically requires the order of words in the sentence.
\begin{equation}
BP = \left\{\begin{matrix}
 1 & \mathrm{if \quad r > c} & \\ 
 e^{(1-r/c)} & \mathrm{if \quad r \leq  c} & 
\end{matrix}\right.
\end{equation}
\begin{equation}
BLEU = \textup{BP} \cdot \textup{exp}\left ( \sum_{n=1}^{N} w_{n} \log p_{n} \right )
\end{equation}
Where $BP$ stands for Brevity Penalty. $c$, $r$ is the length of the predictions and labels, respectively. BP will penalty cases where the model failed to propose correction, or the change happens more than allowed (as the number of words need to be corrected must be equal to the actual corrected). $p_{n}$ stands for modified n-gram precision, using n-grams up to length N and positive weights $w_{n}$ summing to one. The n-gram precision can be simply understands as 'the number of corrected words which occur in reference sentence (ground-truth)' divided by 'the number of words after sentence transformed'. Therefore, the BLEU metrics has potential to not only to keep track strictly of word ordering by measuring n-gram (up-to-4) overlapping but also evaluate how a sentence has been corrected from the original despite the action (remove, edit, add more words).

\subsection{Model Settings}
Our models are implemented by fairseq toolkit \cite{DBLP:journals/corr/abs-1904-01038} which is an re-implementation on the base Transformer architecture \cite{vaswani2017attention}. To find the appropriate hyperparameters for the proposed model, experiments with multiple model designations has been reviewed and the configuration of Jinhua work \cite{zhu2020incorporating} are selected.
Training details with hyperparameter settings are in the table \ref{table:Hyperparameters}:

\FloatBarrier
\begin{table}[h]
	\centering
	\caption{Hyperparameters of Transformer model}
    \begin{tabularx}{\columnwidth}{X >{\raggedleft\arraybackslash}X}
    \hline 
    \\
    BERT model & bert-base-multilingual-cased \\
              & vinai/phobert-base \\
    Number of epochs & 100 \\
    Dropout & 0.3 \\
    Loss Function & labeled smoothed cross-entropy \\
    Optimizer & Adam(0.9,0.98)\\
    Learning Rate & 0.005 \\
    Label Smoothing & 0.1 \\
    Weight Decay & 0.0001 \\
    Beam Search & 5 \\
    Max tokens & 1280 \\
    \\
\end{tabularx}
\label{table:Hyperparameters}
\end{table}
\FloatBarrier

\subsection{Experimental Results and Discussion}
In this phase, we compared with the Google Docs spell checking tool \footnote{The tool can be found on the Google Docs website (https://docs.google.com/). We collected samples by using a web browser behavior simulator based on Selenium framework that manipulate the Google spell checking tool to correct all of its possible suggestions. } and other methods. From the results showed in Figure \ref{table:EvaluationResult}, two versions of our model, Transformer+vinai/ phoBERT and Transformer+BERT-multi-cased, achieved better results than the previous methods. This partly reinforces our hypothesis that using a pre-trained language model BERT brings two benefits to the spelling correction problem: being applicable in the spelling correction task and taking advantage of contextualized word embeddings. Firstly, as mentioned in the BERT paper, tasks such as Text Classification, Question and Answering, Sentence Tagging, etc, are recommended to be used in the fined-tuning phase but the spelling correction task. Due to our modification, at the first step, BERT is verified to be beneficial for the correction task. Secondly, when correcting an error word, the action of choosing a suitable candidate based on context words is the main characteristic of the spelling correction problem. Concretely, BERT produces contextualized word embedding (the same word for different contexts have different embeddings) helps the models to better utilize word embedding at correcting phase. Besides, the pre-training of BERT on a huge data set also makes fine-tuning for our model easier because of no need to re-train from the beginning, taking advantage of the knowledge from the language model.

\FloatBarrier
\begin{table}[h]
    \centering
    \caption{BLEU scores on models}
    \begin{tabularx}{\columnwidth}{|X|X|}
        \hline
        \textbf{Model} & \textbf{BLEU score} \\
        \hline
        Google Docs spellchecking tool & 0.6829 \\
        \hline
        Transformer + vinai/phobert-base & 0.8027 \\
        \hline
        Word2Vec & 0.8222 \\
        \hline
        \textbf{Transformer + bert-multi-cased}  & \textbf{0.8624} \\ 
        \hline
        \multicolumn{2}{|>{\arraybackslash}p{11.8cm}|}{ \footnotesize \textbf{Transformer + vinai/phobert-base}: The proposed model based on the Transformer architecture and PhoBERT \cite{nguyen2020phobert}. } \\
        \multicolumn{2}{|>{\arraybackslash}p{11.8cm}|}{ \footnotesize \textbf{Word2Vec}: The reimplementation of the Word2Vec approach in spelling correction \cite{DBLP:journals/corr/abs-1710-07045}.} \\
        \multicolumn{2}{|>{\arraybackslash}p{11.8cm}|}{ \footnotesize \textbf{Transformer + bert-multi-cased}: The proposed model based on the Transformer architecture and BERT multilingual model \cite{DBLP:journals/corr/abs-1810-04805}.} \\
        \hline
    \end{tabularx}
    \label{table:EvaluationResult}
\end{table}
\FloatBarrier

For the objective of comparison and practical application, there are a few patterns that our excellent model gain out performance: telex and edit-distance error types, compared to the google docs spellchecking tool. This happened partially because we designated more of these types of error to achieve our goal. More tuning is needed in future work on the error type distribution to improve performance for other types of errors.

The google docs spellchecking tool has another advantage over our model is the ability to restrict unnecessary correction. Additionally, the emergence of proper nouns also makes our model ineffective. When it comes to a proper noun, especially Vietnamese proper names, our model tends to correct them, which should not be the case. To overcome this weakness, some supporting components can be developed to the proposed architecture: Applying a name entity recognition component or an independent spellchecker to determine to correct a word or not.

\section{Conclusion}\label{Conclusion}
In this paper, a combination of BERT and Transformer architecture is implemented for the Vietnamese spell correction task. The experimental results show that our model outperforms other approaches with a 0.86 BLUE score and can be used in real-world applications. Besides, a dataset is constructed for related works based on a breakdown of the spelling correction problem to define which errors commonly happened and need more attention.

To our concern, despite the improvement in the model's performance, there are late inferences due to large and complex architecture. In addition, due to the different distribution of data in the pre-trained model compared to data for the spelling correction task, we can not fully utilize the representation of pre-trained words, resulting in the model sometimes try to correct the unwanted words.

In the future, our architecture will be experimented with other existing pre-trained language models to see how well the compatibility they are. Moreover, we also evaluate our model's accuracy on a bigger dataset. Finally, investigating and analyzing errors that may happen in practices is our priority in order to create a better error pseudo generator. 

%
%
%
%

\bibliographystyle{splncs04}
\bibliography{reference}

\end{document}